\documentclass[runningheads]{llncs}
\usepackage[T1]{fontenc}
\usepackage{graphicx}
\usepackage{amssymb}
\usepackage{amsmath}
\usepackage{booktabs}
\usepackage[misc]{ifsym}
\begin{document}
\title{A Spatial-Temporal Deformable Attention based Framework for Breast Lesion Detection in Videos}
%
%
\authorrunning{C. Qin et al.}
\author{Chao Qin\inst{1}$^{\textrm{(\Letter)}}$, Jiale Cao\inst{2}, Huazhu Fu\inst{3}, Rao Muhammad Anwer\inst{1}, \\ Fahad Shahbaz Khan\inst{1, 4}}  

\institute{Mohamed bin Zayed University of Artificial Intelligence, Abu Dhabi, United Arab Emirates \\ \email{chao.qin@mbzuai.ac.ae} \\ \and Tianjin University, Tianjin, China \and Institute of High Performance Computing, Agency for Science, Technology and Research, Singapore \and 
Linköping University, Linköping, Sweden}
%
\maketitle              
\begin{abstract}
Detecting breast lesion in videos is crucial for computer-aided diagnosis. Existing video-based breast lesion detection approaches typically perform temporal feature aggregation of deep backbone features based on the self-attention operation. We argue that such a strategy struggles to effectively perform deep feature aggregation and ignores the useful local information. To tackle these issues, we propose a spatial-temporal deformable attention based framework, named STNet. Our STNet introduces a spatial-temporal deformable attention module to perform local spatial-temporal feature fusion. The spatial-temporal deformable attention module enables deep feature aggregation in each stage of both encoder and decoder. To further accelerate the detection speed, we introduce an encoder feature shuffle strategy for multi-frame prediction during inference. In our encoder feature shuffle strategy, we share the backbone and encoder features, and shuffle encoder features for decoder to generate the predictions of multiple frames. The experiments on the public breast lesion ultrasound video dataset show that our STNet obtains a state-of-the-art detection performance, while operating twice as fast inference speed. The code and  model are available at \url{https://github.com/AlfredQin/STNet}.


\end{abstract}
\section{Introduction}
Ultrasound imaging is a very effective technique for breast lesion diagnosis, which has high sensitivity. Automatically detecting breast lesions is a challenging problem with a potential to aid in improving the efficiency of radiologists in ultrasound-based breast cancer diagnosis \cite{yap2017automated,zhu2020second}. Some of the challenges associated with automatic breast lesion detection include blurry boundaries and changeable sizes of breast lesions.


Most existing breast lesion detection methods can  be categorized into image-based \cite{zhang2020birads,movahedi2020automated,qi2019automated,xue2021global,yang2020temporal}  and video-based \cite{Chen_SSB_arxiv_2019,lin2022new} breast lesion detection approaches. Image-based breast lesion detection approaches perform detection in each frame independently. Compared to image-based breast lesion detection approaches, methods based on videos are capable of utilizing temporal information for improved detection performance. For instance, Chen \textit{et al.} \cite{Chen_SSB_arxiv_2019} exploited temporal coherence for semi-supervised video-based breast lesion detection. Recently, Lin \textit{et al.} \cite{lin2022new} proposed a feature aggregation network, termed as CVA-Net, that executes intra-video and inter-video fusions at both video and clip levels based on attention blocks. Although the recent CVA-Net aggregates clip and video level features, we distinguish two key issues that hamper its performance. 
First,  the self-attention based cross-frame feature fusion is a global-level operation and it operates once before the encoder-decoder, thereby ignoring the useful local information and in turn missing an effective deep feature fusion.  Second, CVA-Net only performs one-frame prediction based on multiple frame inputs, which is very time-consuming.
 
To address the aforementioned issues, we propose a spatial-temporal deformable attention based network, named STNet, for detecting the breast lesions in ultrasound videos. Within our STNet, we introduce a spatial-temporal deformable attention module to fuse multi-scale spatial-temporal information among different frames, and further integrate it into each layer of the encoder and decoder. In this way, different from the recent CVA-Net, our proposed STNet performs both deep and local feature fusion. In addition, we introduce multi-frame prediction with encoder feature shuffle operation that shares the backbone and encoder features, and only perform multi-frame prediction in the decoder. This enables us to significantly accelerate the detection speed of the proposed approach. We conduct extensive experiments on a public breast lesion ultrasound video dataset, named BLUVD-186 \cite{lin2022new}. The experimental results validate the efficacy of our proposed STNet that has a superior detection performance. For example, our proposed STNet  achieves a mAP of 40.0\% with an absolute gain of 3.9\% in terms of detection accuracy, while operating at two times faster, compared to the recent CVA-Net \cite{lin2022new}.

%

\section{Method}
\begin{figure}[t!]
\includegraphics[width=1.0\linewidth]{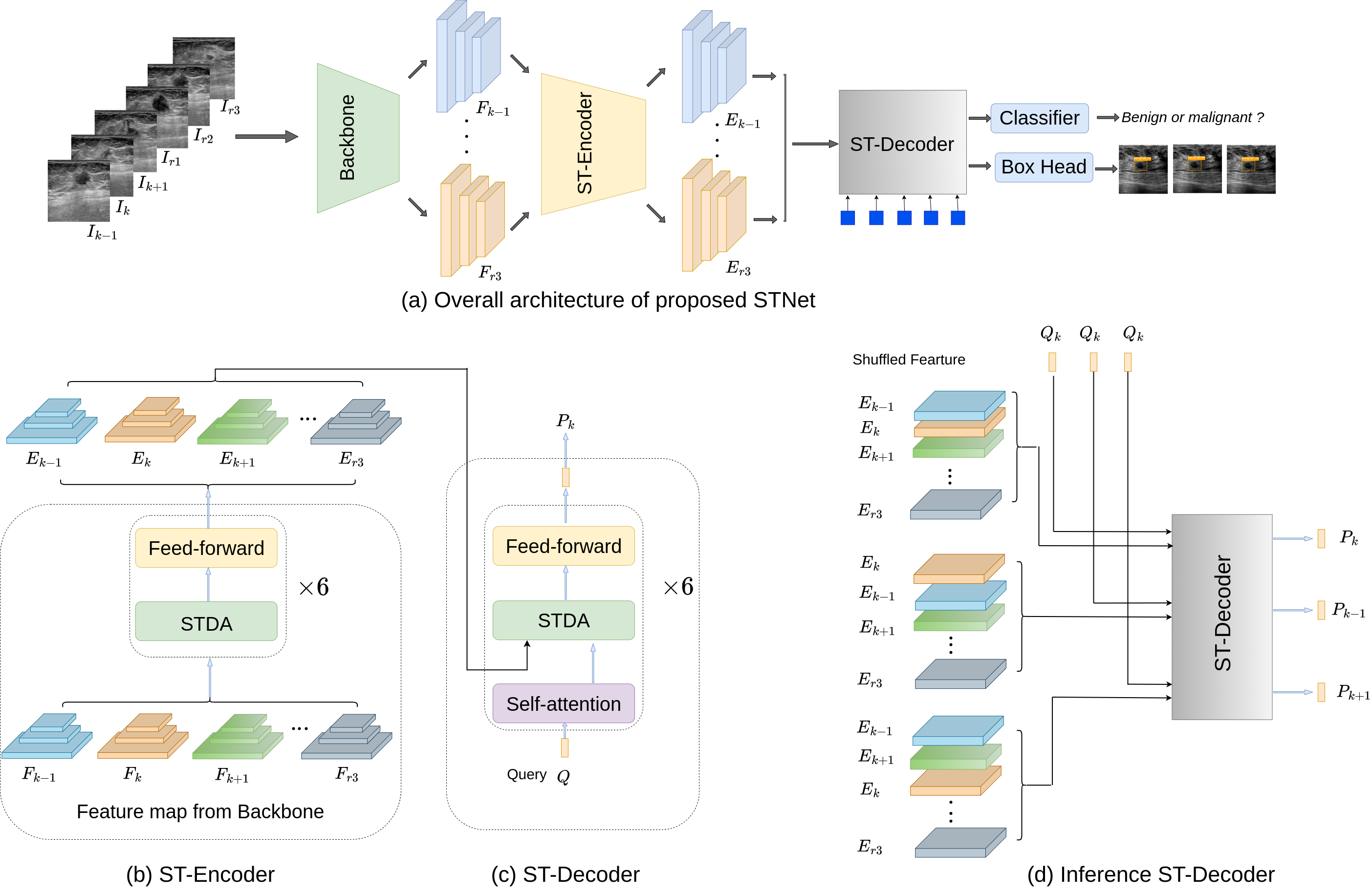} 
\caption{\textbf{(a)} Overall architecture of the proposed STNet. The proposed STNet takes six frames as inputs and extracts multi-scale features of each frame. Afterwards, the proposed STNet utilizes a spatial-temporal deformable attention (STDA) based encoder \textbf{(b)} and decoder \textbf{(c)} for spatial-temporal multi-scale information fusion. Finally, the proposed STNet performs classification and regression. \textbf{(d)} During inference, we introduce a encoder feature shuffle strategy for multi-frame prediction.} 
\label{fig:model}
\end{figure}

Here, we describe our proposed spatial-temporal deformable attention based framework, named STNet, for detecting breast lesions in the ultrasound videos. 
Fig.\ref{fig:model}(a) presents the overall architecture of our proposed STNet, which is built on the end-to-end detector deformable DETR  \cite{Zhu_DeformableDETR_ICLR_2021}. Within our STNet, we introduce spatial-temporal deformable attention into the encoder and the decoder. As in CVA-Net \cite{lin2022new},  we take  six frames $I_{k-1}, I_k$, $I_{k+1}$, $I_{r1}$, $I_{r2}$, $I_{r3}$ from one ultrasound video as inputs, where there are three neighboring frames $I_{k-1}, I_k$, $I_{k+1}$ and three randomly-selected frames $I_{r1}$, $I_{r2}$, $I_{r3}$. Given these input frames, we use the  backbone, such as ResNet-50 \cite{He_ResNet_CVPR_2016}, to extract deep multi-scale features  $F_{k-1}, F_k$, $F_{k+1}$, $F_{r1}$, $F_{r2}$, $F_{r3}$. Afterwards, we introduce a spatial-temporal deformable attention based encoder (ST-Encoder) to perform intra-frame and inter-frame multi-scale feature fusion. Then, we introduce a spatial-temporal deformable attention based decoder  (ST-Decoder) to generate output feature embeddings $P_k$, which are fed to a classifier and a box predictor for classification and bounding-box regression. During inference, we take three neighboring frames  and three randomly-selected frames as the inputs, and  simultaneously predict the results of three neighboring frames using our encoder feature shuffle strategy. As a result, our approach operates at a faster inference speed.


\subsection{Spatial-Temporal Deformable Attention}
\begin{figure}[t!]
\includegraphics[width=1.0\linewidth]{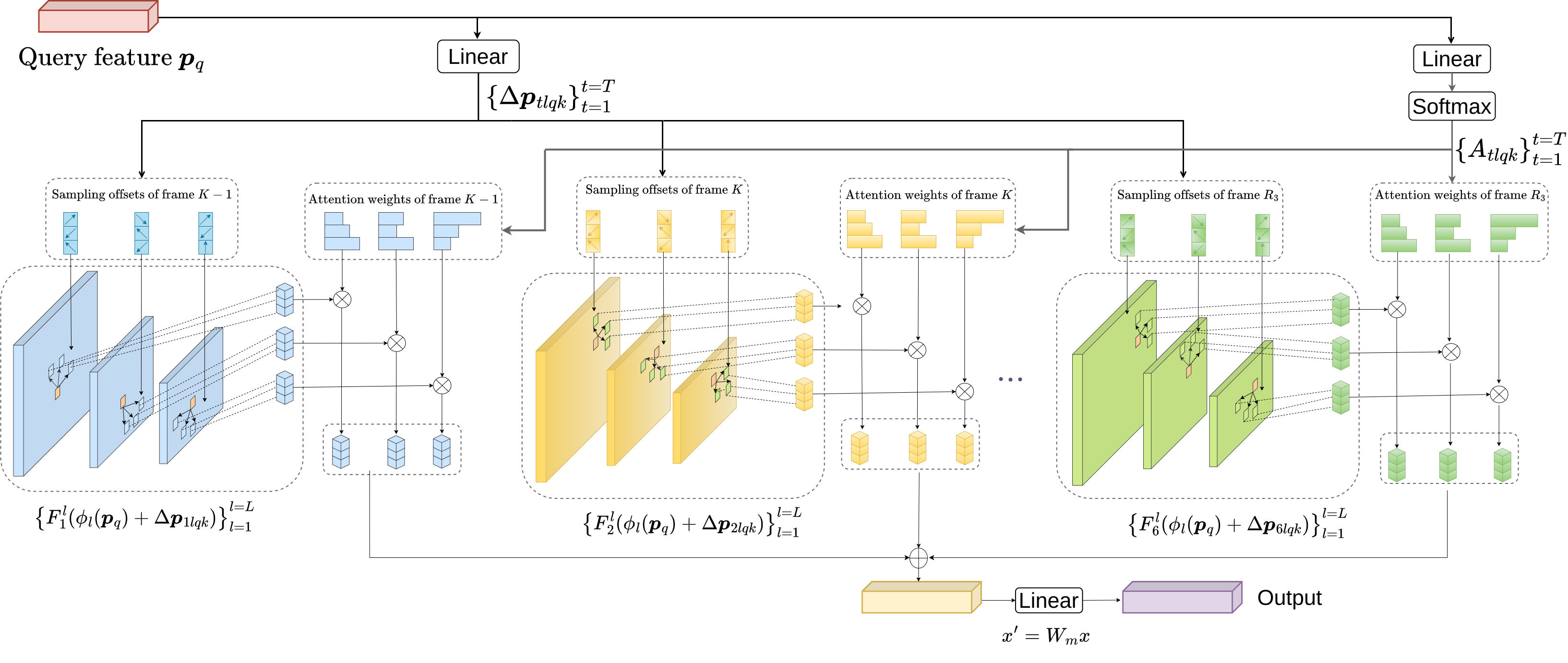} 
\caption{Structure of our proposed Spatial-temporal deformable attention (STDA). Given a query feature and reference point, our STDA not only fuses multi-scale features within a frame, but also aggregates multi-scale features between different frames.} 
\label{fig:stda}
\end{figure}

Given a reference point, deformable attention \cite{Zhu_DeformableDETR_ICLR_2021} aggregates  the features of a group of key sampling points near it. Compared to original transformer self-attention  \cite{Vaswani_Att_NIPS_2017}, deformable attention has low-complexity along with a faster convergence speed. Motivated by this, we adopt deformable attention for breast lesion detection and extend it to spatial-temporal deformable attention (STDA). Our STDA not only aggregates the features of current frame, but also aggregates the features of the rest of the frames. Fig. \ref{fig:stda} presents the  structure of our proposed STDA.  Let $F_t=\left \{ F^l_t \right \} ^L_{l=1}$ represent the set of multi-scale feature maps  at frame $t$, where $F_t^l \in \mathbb{R}^{C\times H_l\times W_l}$ is the feature map at level $l$. Given the  query features $\boldsymbol{p}_{q}$ and corresponding reference points $\boldsymbol{z}_{q}$, the spatio-temporal multi-scale attention is given as:
\begin{equation}
\operatorname{STDA}\left(\boldsymbol{z}_{q}, \boldsymbol{p}_{q}, \left \{F_t \right \} ^{T }_{t=0}\right) = \sum_{m=1}^{M} W_m \sum_{t=1}^{T} \sum_{l=1}^{L} \sum_{k=1}^{K} A_{t l q k} F_t^l(\phi_{l}(\boldsymbol{p}_{q})+\Delta \boldsymbol{p}_{t l q k}),
\end{equation}
where $m$ represents multi-head index and $k$ is sampling point index. $W_m$ is a linear layer, $A_{t l q k}$ indicates  attention weight of  sampling point, and  $\Delta \boldsymbol{p}_{t l q k}$ indicates sample offset of  sampling point.  $\phi_{l}$ normalizes the coordinates $\boldsymbol{p}_{q}$ by the scale of feature map $F_t^l$. The sampling offset $\Delta \boldsymbol{p}_{t l q k}$  is predicted by the query feature $\boldsymbol{z}_{q}$ with a linear layer. The attention weight $A_{t l q k}$ is predicted by feeding query feature $\boldsymbol{z}_{q}$ to a linear layer and a softmax layer. As a result,  the sum of attention weights is equal to one as 
\begin{equation}
\sum_{t=1}^{T} \sum_{l=1}^{L} \sum_{k=1}^{K} A_{t l q k}=1.
\end{equation}
Compared to the standard deformable attention, the proposed spatial-temporal deformable attention fully exploits spatial information within frame and temporal information across frames.

\subsection{Spatial-Temporal Deformable Attention based Encoder and Decoder}
Here, we integrate the proposed spatial-temporal deformable attention (STDA) into encoder and decoder (called ST-Encoder and ST-Decoder ). As shown in Fig. \ref{fig:model}(b), ST-Encoder takes deep multi-scale feature maps $F_{k-1}, F_k$, $F_{k+1}$, $F_{r1}$, $F_{r2}$, $F_{r3}$ as inputs. Afterwards, we employ STDA to perform spatial and temporal fusion and generate the fused multi-scale feature maps $F'_{k-1}, F'_k$, $F'_{k+1}$, $F'_{r1}$, $F'_{r2}$, $F'_{r3}$, where the query  corresponds to each pixel  in multi-scale feature maps. Then, the fused feature map goes through a feed-forward network (FFN) to generate the output feature maps $E_{k-1}, E_k$, $E_{k+1}$, $E_{r1}$, $E_{r2}$, $E_{r3}$. Similar to the original deformable DETR, we adopt cascade structure to stack six STDA and FFN layers in ST-Encoder.

The ST-Decoder takes the output feature maps $E_{k-1}, E_k$, $E_{k+1}$, $E_{r1}$, $E_{r2}$, $E_{r3}$ and a set of learnable queries $Q\in \mathbb{R}^{N\times C}$ as inputs. The learnable queries first go through a self-attention layer. Afterwards, STDA performs cross-attention operation between these feature maps and the queries, where the key elements are these output feature maps of ST-Encoder. Then, we employ a FFN layer to generate the prediction features $P_k\in \mathbb{R}^{N\times C}$. We also stack six self-attention, STDA, and FFN  layers in ST-Decoder for deep feature extraction.

\subsection{Multi-frame Prediction With Encoder Feature Shuffle}
As discussed above, the proposed STNet adopts six frames to predict the results of one frame. Although STNet fully exploits temporal information for improved breast lesion detection, it becomes time-consuming for multi-frame prediction. To accelerate the detection speed, we introduce multi-frame prediction with encoder feature shuffle during inference. Instead of going through the entire network several times, we first share deep multi-scale feature maps before encoder and second perform the decoder several times for multi-frame prediction. To perform multi-frame prediction only in the decoder, we propose the encoder feature shuffle operation shown in Fig. \ref{fig:model}(d). By exchanging the order of neighboring frame $I_{k-1}, I_{k}, I_{k+1}$, the decoder can predict the results of three neighboring frames, respectively. Compared to the original STNet, the proposed encoder feature shuffle strategy only employs decoder forward three frames and accelerates the inference speed.

\section{Experiments}
\subsection{Dataset and Implementation Details}
\textbf{Dataset} We conduct the experiments on the public BLUVD-186 dataset \cite{lin2022new}, comprising 186 videos including 112 malignant and 74 benign cases. The dataset has totally 25,458 ultrasound frames, where the number of frames in a video ranges from 28 to 413. The videos encompass a comprehensive tumor scan, from its initial appearance to its largest section and eventual disappearance. All videos were captured using PHILIPS TIS L9-3 and LOGIQ-E9. 
The grounding-truths in a frame, including breast lesion bounding-boxes and corresponding categories, are labeled by two pathologists, which have eight years of professional background in the field of breast pathology.
We adopt the same dataset splits as in the previous work CVA-Net \cite{lin2022new}, to guarantee a fair comparison. Specifically, the testing set comprises 38 videos  randomly selected from all 186 videos, while the rest of the videos are used as the training set.

\noindent \textbf{Evaluation Metrics} Three commonly-used metrics are employed for performance evaluation of breast lesion detection methods on the ultrasound videos, namely average precision (AP), $\mathrm{AP}_{50}$, and $\mathrm{AP}_{75}$.

\noindent \textbf{Implementation Details} We employ the ResNet-50 \cite{He_ResNet_CVPR_2016} pre-trained on ImageNet \cite{deng2009imagenet}, and use Xavier \cite{glorot2010understanding} to initialize the remaining network parameters. To enhance the diversity of training data, all videos are randomly subjected to horizontal flipping, cropping, and resizing. Similar to that of CVA-Net, we employ a two-phase training strategy to  achieve better convergence. In the first phase, we employ Adam optimizer to train the model for 8 epochs. We then fine-tune the model for another 20 epochs with the SGD optimizer. Throughout both phases of training, we adopt the consistent hyper-parameters, where the learning rate is $5 \times 10^{-5}$ and the weight decay is $1\times10^{-4}$. We train the model on a single NVIDIA A100 GPU and set the batch size as 1. 

\begin{table}[!t]
\caption{
State-of-the-art quantitative comparison of our approach with existing methods in literature on the BLUVD-186 dataset. Our approach achieves a superior performance on three different metrics. Compared to the recent CVA-Net~\cite{lin2022new}, our approach obtains a gain of 3.9\% in terms of overall AP. We show the best results in bold.
}
\label{tab:sota}
\centering
\vspace{-3mm}
\begin{tabular}{l|c|c|ccc}
\bottomrule
  Method & Type & Backbone & AP & AP$_{50}$ & AP$_{75}$ \\
\hline
  GFL~\cite{li2020generalized} & image & ResNet-50 & 23.4 & 46.3 & 22.2 \\
  Cascade RPN~\cite{vu2019cascade} & image & ResNet-50 & 24.8 & 42.4 & 27.3\\
  Faster R-CNN~\cite{ren2015faster} & image & ResNet-50 & 25.2 & 49.2 & 22.3 \\
  VFNet~\cite{Zhang2021VarifocalNetAI} & image & ResNet-50 & 28.0 & 47.1 & 31.0 \\
  RetinaNet~\cite{lin2017focal} & image & ResNet-50 & 29.5 & 50.4 & 32.4 \\
\hline
  DFF~\cite{zhu2017deep} & video & ResNet-50 & 25.8 & 48.5 & 25.1 \\
  FGFA~\cite{zhu2017flow} & video & ResNet-50 & 26.1 & 49.7 & 27.0 \\
  SELSA~\cite{wu2019sequence} & video & ResNet-50 & 26.4 & 45.6 & 29.6 \\
  Temporal ROI Align~\cite{gong2021temporal} & video & ResNet-50 & 29.0 & 49.9 & 33.1 \\
  MEGA~\cite{chen2020memory} & video & ResNet-50 & 32.3 & 57.2 & 35.7 \\
  CVA-Net~\cite{lin2022new} & video & ResNet-50 & 36.1 & 65.1 & 38.5 \\
\hline
  \textbf{STNet (Ours)} & video & ResNet-50 & \textbf{40.0} & \textbf{70.3} & \textbf{43.3} \\
\toprule
\end{tabular}
\end{table}

\subsection{State-of-the-Art Comparison}
Our proposed approach is compared with eleven state-of-the-art methods, comprising  image-based and video-based methods. 
We report the detection performance of these state-of-the-art methods generated by CVA-Net \cite{Zhi_CVA_MICCAI_2022}. Specifically,  CVA-Net acquires the detection performance of these  methods by utilizing their publicly available codes or re-implementing them if no publicly available codes.

\noindent \textbf{Quantitative Comparisons} 
Table \ref{tab:sota} presents the state-of-the-art quantitative comparison of our approach with the eleven existing breast lesion video detection methods in literature. As a general trend, video-based methods tend to yield higher average precision (AP), AP50, and AP75 scores compared to image-based breast lesion detection methods. Among the eleven existing methods, the recent CVA-Net~\cite{lin2022new} achieves the best overall AP score of 36.1, AP50 score of 65.1, and AP75 score of 38.5. Our proposed STNet method consistently outperforms CVA-Net~\cite{lin2022new} on all three metrics (AP, AP50, and AP75). Specifically, our STNet achieves a significant improvement in the overall AP score from 36.1 to 40.0, the AP50 score from 65.1 to 70.3, and the AP75 score from 38.5 to 43.3. The significant improvement demonstrates the efficacy of our approach for detecting breast lesions in ultrasound videos.

\begin{figure}[t!]
\includegraphics[width=1.0\linewidth]{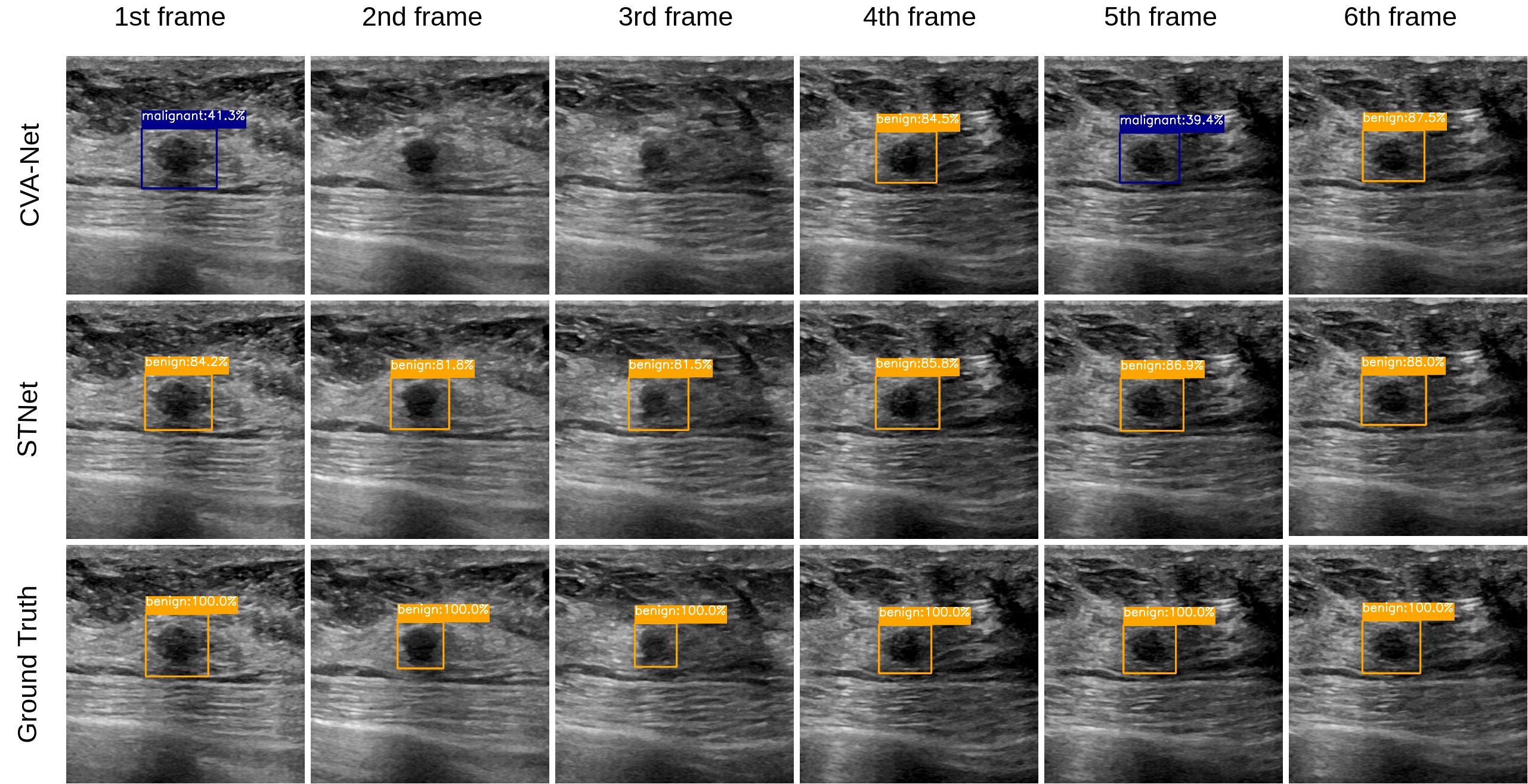} 
\caption{Qualitative breast lesion detection comparison on example ultrasound video frames between the recent CVA-Net \cite{lin2022new} and our proposed STNet. We also show the ground truth as reference. Our STNet achieves improved detection performance, compared to CVA-Net. Best viewed zoomed in.
} 
\label{fig:comparision}
\end{figure}

\noindent \textbf{Qualitative Comparisons} 
Fig.~\ref{fig:comparision} presents the qualitative breast lesion detection comparison  between CVA-Net and our proposed approach on an ultrasound video containing the benign breast lesions. Moreover, we show the ground truth of each frame on the third row for reference. The first row of the figure shows that CVA-Net struggles to identify the breast lesions in the second and third frames. Further, although CVA-Net manages to identify the breast lesions in the first and fifth frames, the classification results are inaccurate (as highlighted by the blue rectangle in Fig. \ref{fig:comparision}). In contrast, our STNet method in the second row of Fig. \ref{fig:comparision} accurately detects the breast lesions in all video frames and achieves accurate classification performance for each frame.

\noindent \textbf{Inference Speed Comparison}
We present the inference speed comparison between our proposed STNet and CVA-Net on an NVIDIA RTX 3090 GPU using the same environment. We use FPS (frames per second) as the performance metric. Specifically, our proposed STNet  achieves an averaged inference speed of 21.84 FPS, while CVA-Net achieves an averaged speed of 12.17 FPS. Our model operates around two times faster than CVA-Net, which we attribute to the ability of our model to predict three frames simultaneously.

\begin{table}[!t]
\caption{
  Ablation study with different design choices. Our proposed STNet achieves a superior performance compared to the baseline and some different designs. We show the est results in bold.
}
\label{tab:ablation}
\centering
\begin{tabular}{l|ccc}
\bottomrule
 & AP & AP$_{50}$ & AP$_{75}$ \\
\hline
  Baseline + Single-frame  & 30.2 & 55.0 & 31.7 \\
  Baseline + Multi-frame & 35.1 & 61.6 & 37.4 \\
  ST-Encoder + DA-Decoder & 34.9  & 59.8  & 37.7  \\
  DA-Encoder + ST-Decoder & 35.8  & 60.4 & 38.0 \\
\hline
  \textbf{STNet (Ours)} & \textbf{40.0} & \textbf{70.3} & \textbf{43.3} \\
\toprule
\end{tabular}
\end{table}

\subsection{Ablation Study}
\noindent \textbf{Effectiveness of STDA:} To show the efficacy of our proposed STDA, we perform  different ablation studies. The first baseline network, referred as "Baseline + Single-frame", uses the original deformable DETR and takes a single frame as input. The second baseline network, referred as "Baseline + Multi-frame", uses modified deformable DETR with multi-head attention module to  fuse six input frames. For the third study, labeled "ST-Encoder + DA-Decoder", we retain the encoder with STDA in our model but replace the STDA in the decoder with the conventional deformable attention. Similarly, in the fourth study, labeled "DA-Encoder + ST-Decoder", we retain the decoder with STDA in our model but replace the STDA in the encoder with the conventional deformable attention. As shown in Table \ref{tab:ablation}, the results show that "ST-Encoder + DA-Decoder" and "DA-Encoder + ST-Decoder" improve the AP by 4.7 and 5.6, respectively, compared to "Baseline + Single-frame". This demonstrates that STDA can effectively perform intra-frame and inter-frame multi-scale feature fusion, even when only partially adopted in the encoder or decoder. Furthermore, our proposed STNet improves the AP by 5.1 and 4.2 compared to "ST-Encoder + DA-Decoder" and "DA-Encoder + ST-Decoder", respectively, indicating that the integration of STDA in both the encoder and decoder is crucial for achieving superior detection performance. 
\section{Conclusion}
 We propose a novel breast lesion detection approach for ultrasound videos, termed as STNet, which performs local spatial-temporal feature fusion and deep feature aggregation in each stage of both encoder and decoder using our spatial-temporal deformable attention module. Additionally, we introduce the encoder feature shuffle strategy that enables multi-frame prediction during inference, thereby enabling us to accelerate the inference speed while maintaining better detection performance. The experiments conducted on a public breast lesion ultrasound video dataset show the efficacy of our  STNet, resulting in a superior detection performance  while operating at a fast inference speed. We believe STNet presents a promising solution and will help further promote future research in the direction of efficient and accurate breast lesion detection in videos.

\vspace{4mm}
\noindent
\textbf{Acknowledgment.} This research is supported by the National Research Foundation, Singapore under its AI Singapore Programme (AISG Award No: AISG2-TC-2021-003) and Agency for Science, Technology and Research (A*STAR) Central Research Fund (CRF).

\bibliographystyle{splncs04}
\bibliography{paper1755}

\begin{thebibliography}{10}
\providecommand{\url}[1]{\texttt{#1}}
\providecommand{\urlprefix}{URL }
\providecommand{\doi}[1]{https://doi.org/#1}

\bibitem{Chen_SSB_arxiv_2019}
Chen, S., Yu, W., Ma, K., Sun, X., Lin, X., Sun, D., Zheng, Y.: Semi-supervised breast lesion detection in ultrasound video based on temporal coherence. In: arXiv:1907.06941 (2019)

\bibitem{chen2020memory}
Chen, Y., Cao, Y., Hu, H., Wang, L.: Memory enhanced global-local aggregation for video object detection. In: Proceedings of the IEEE/CVF Conference on Computer Vision and Pattern Recognition. pp. 10337--10346 (2020)

\bibitem{deng2009imagenet}
Deng, J., Dong, W., Socher, R., Li, L.J., Li, K., Fei-Fei, L.: Imagenet: A large-scale hierarchical image database. In: 2009 IEEE conference on computer vision and pattern recognition. pp. 248--255. Ieee (2009)

\bibitem{glorot2010understanding}
Glorot, X., Bengio, Y.: Understanding the difficulty of training deep feedforward neural networks. In: Proceedings of the thirteenth international conference on artificial intelligence and statistics. pp. 249--256. JMLR Workshop and Conference Proceedings (2010)

\bibitem{gong2021temporal}
Gong, T., Chen, K., Wang, X., Chu, Q., Zhu, F., Lin, D., Yu, N., Feng, H.: Temporal roi align for video object recognition. In: Proceedings of the AAAI Conference on Artificial Intelligence. vol.~35, pp. 1442--1450 (2021)

\bibitem{He_ResNet_CVPR_2016}
He, K., Zhang, X., Ren, S., Sun, J.: Deep residual learning for image recognition. In: IEEE Conference on Computer Vision and Pattern Recognition (2016)

\bibitem{li2020generalized}
Li, X., Wang, W., Wu, L., Chen, S., Hu, X., Li, J., Tang, J., Yang, J.: Generalized focal loss: Learning qualified and distributed bounding boxes for dense object detection. Advances in Neural Information Processing Systems  \textbf{33},  21002--21012 (2020)

\bibitem{lin2017focal}
Lin, T.Y., Goyal, P., Girshick, R., He, K., Doll{\'a}r, P.: Focal loss for dense object detection. In: Proceedings of the IEEE International Conference on Computer Vision. pp. 2980--2988 (2017)

\bibitem{lin2022new}
Lin, Z., Lin, J., Zhu, L., Fu, H., Qin, J., Wang, L.: A new dataset and a baseline model for breast lesion detection in ultrasound videos. In: Medical Image Computing and Computer Assisted Intervention--MICCAI 2022: 25th International Conference, Singapore, September 18--22, 2022, Proceedings, Part III. pp. 614--623. Springer (2022)

\bibitem{Zhi_CVA_MICCAI_2022}
Lin, Z., Lin, J., Zhu, L., Fu, H., Qin, J., Wang, L.: A new dataset and a baseline model for breast lesion detection in ultrasound videos. In: Medical Image Computing and Computer Assisted Intervention. pp. 614--623 (2022)

\bibitem{movahedi2020automated}
Movahedi, M.M., Zamani, A., Parsaei, H., Tavakoli~Golpaygani, A., Haghighi~Poya, M.R.: Automated analysis of ultrasound videos for detection of breast lesions. Middle East Journal of Cancer  \textbf{11}(1),  80--90 (2020)

\bibitem{qi2019automated}
Qi, X., Zhang, L., Chen, Y., Pi, Y., Chen, Y., Lv, Q., Yi, Z.: Automated diagnosis of breast ultrasonography images using deep neural networks. Medical image analysis  \textbf{52},  185--198 (2019)

\bibitem{ren2015faster}
Ren, S., He, K., Girshick, R., Sun, J.: Faster r-cnn: Towards real-time object detection with region proposal networks. arXiv preprint arXiv:1506.01497  (2015)

\bibitem{Vaswani_Att_NIPS_2017}
Vaswani, A., Shazeer, N., Parmar, N., Uszkoreit, J., Jones, L., Gomez, A.N., Kaiser, L., Polosukhin, I.: Attention is all you need. In: Advances in Neural Information Processing Systems (2017)

\bibitem{vu2019cascade}
Vu, T., Jang, H., Pham, T.X., Yoo, C.D.: Cascade rpn: Delving into high-quality region proposal network with adaptive convolution. arXiv preprint arXiv:1909.06720  (2019)

\bibitem{wu2019sequence}
Wu, H., Chen, Y., Wang, N., Zhang, Z.: Sequence level semantics aggregation for video object detection. In: Proceedings of the IEEE/CVF International Conference on Computer Vision. pp. 9217--9225 (2019)

\bibitem{xue2021global}
Xue, C., Zhu, L., Fu, H., Hu, X., Li, X., Zhang, H., Heng, P.A.: Global guidance network for breast lesion segmentation in ultrasound images. Medical image analysis  \textbf{70},  101989 (2021)

\bibitem{yang2020temporal}
Yang, Z., Gong, X., Guo, Y., Liu, W.: A temporal sequence dual-branch network for classifying hybrid ultrasound data of breast cancer. IEEE Access  \textbf{8},  82688--82699 (2020)

\bibitem{yap2017automated}
Yap, M.H., Pons, G., Marti, J., Ganau, S., Sentis, M., Zwiggelaar, R., Davison, A.K., Marti, R.: Automated breast ultrasound lesions detection using convolutional neural networks. IEEE journal of biomedical and health informatics  \textbf{22}(4),  1218--1226 (2017)

\bibitem{zhang2020birads}
Zhang, E., Seiler, S., Chen, M., Lu, W., Gu, X.: Birads features-oriented semi-supervised deep learning for breast ultrasound computer-aided diagnosis. Physics in Medicine \& Biology  \textbf{65}(12),  125005 (2020)

\bibitem{Zhang2021VarifocalNetAI}
Zhang, H., Wang, Y., Dayoub, F., Sunderhauf, N.: Varifocalnet: An iou-aware dense object detector. IEEE/CVF Conference on Computer Vision and Pattern Recognition (CVPR) pp. 8510--8519 (2021)

\bibitem{zhu2020second}
Zhu, L., Chen, R., Fu, H., Xie, C., Wang, L., Wan, L., Heng, P.A.: A second-order subregion pooling network for breast lesion segmentation in ultrasound. In: International Conference on Medical Image Computing and Computer-Assisted Intervention. pp. 160--170. Springer (2020)

\bibitem{Zhu_DeformableDETR_ICLR_2021}
Zhu, X., Su, W., Lu, L., Li, B., Wang, X., Dai, J.: Deformable detr: Deformable transformers for end-to-end object detection. In: International Conference on Learning Representations (2021)

\bibitem{zhu2017flow}
Zhu, X., Wang, Y., Dai, J., Yuan, L., Wei, Y.: Flow-guided feature aggregation for video object detection. In: Proceedings of the IEEE International Conference on Computer Vision. pp. 408--417 (2017)

\bibitem{zhu2017deep}
Zhu, X., Xiong, Y., Dai, J., Yuan, L., Wei, Y.: Deep feature flow for video recognition. In: Proceedings of the IEEE Conference on Computer Vision and Pattern Recognition. pp. 2349--2358 (2017)

\end{thebibliography}




\end{document}